%% file: root.tex
\documentclass{article} % For LaTeX2e
\usepackage{iclr2024_arxiv,times}

\input{math_commands.tex}

\usepackage{hyperref}
\usepackage{url}

\usepackage{tabularx}
\usepackage{graphicx}
\usepackage{arydshln}
\usepackage{amsmath} % assumes amsmath package installed
\usepackage{amssymb}  % assumes amsmath package installed
\usepackage{bm}
\usepackage{comment}
\usepackage{multirow}
\usepackage{booktabs}
\usepackage{makecell}
\usepackage{subfig}

\usepackage[noend]{algpseudocode}% http://ctan.org/pkg/algorithmicx
\usepackage{algorithm}% http://ctan.org/pkg/algorithm

\title{Diff-Transfer: Model-based Robotic Manipulation Skill Transfer via Differentiable Physics Simulation}

\iclrfinalcopy
\author{Yuqi Xiang$^1$, Feitong Chen$^2$, Qinsi Wang$^3$, Yang Gang$^2$, Xiang Zhang$^4$, Xinghao Zhu$^4$, \\ 
\textbf{Xingyu Liu}$^5$, \textbf{Lin Shao}$^2$ \\
$^1$Nanjing University \quad $^2$National University of Singapore \\
$^3$University of Science and Technology of China  \quad  $^4$University of California, Berkeley \\ 
$^5$Carnegie Mellon University
}

\begin{document}

\maketitle

\begin{abstract}
\input{tex/abs}
\end{abstract}

\section{Introduction}

\input{tex/intro}

\section{Related Work}
\input{tex/relatedwork}

\section{Problem Statement} \label{probdef}
\input{tex/prodef}

\section{Technical Approach} \label{sec:tech}
\input{tex/approach}

\section{Experiments}
\input{tex/exp}

\section{Conclusion}
\label{sec:conclusion}
\input{tex/con}

\bibliography{iclr2024_conference}
\bibliographystyle{iclr2024_conference}

% \appendix
% \section{Appendix}

\end{document}

%% file: math_commands.tex
\usepackage{amsmath,amsfonts,bm}

\def\eqref#1{equation~\ref{#1}}
\def\1{\bm{1}}

\DeclareMathAlphabet{\mathsfit}{\encodingdefault}{\sfdefault}{m}{sl}
\SetMathAlphabet{\mathsfit}{bold}{\encodingdefault}{\sfdefault}{bx}{n}

%% file: tex/abs.tex
% In this work, we consider the problem of transferring a policy between two tasks including different object poses and object shapes. We propose

% For intelligent robots operating in the real world, swiftly acquiring skills to accomplish new manipulation tasks with objects is crucial. A naive approach of training a manipulation policy for each object on every task from scratch is inefficient and impractical due to the vast diversity in object shapes, physical properties, and potential tasks. We propose a novel framework, Diff-Transfer, leveraging differentiable physics simulation functions, aiming to address the intricate dynamics of object manipulations. Diff-Transfer tackles this problem by interpolating between two tasks and generating numerous intermediate sub-tasks that progressively transition from the source task to the target task. The principal transfer target task encompasses two scenarios: object pose transfer and object shape transfer. Diff-Transfer resolves this by treating their sub-tasks independently and employing varied methodologies to generate the path from the source task to the target task. We implement our framework in simulation experiments and execute four pose transfer tasks and two shape transfer tasks, demonstrating the efficacy of Diff-Transfer through comprehensive experiments.

The capability to transfer mastered skills to accomplish a range of similar yet novel tasks is crucial for intelligent robots. In this work, we introduce \textit{Diff-Transfer}, a novel framework leveraging differentiable physics simulation to efficiently transfer robotic skills. Specifically, \textit{Diff-Transfer} discovers a feasible path within the task space that brings the source task to the target task. At each pair of adjacent points along this task path, which is two sub-tasks, \textit{Diff-Transfer} adapts known actions from one sub-task to tackle the other sub-task successfully. The adaptation is guided by the gradient information from differentiable physics simulations. We propose a novel path-planning method to generate sub-tasks, leveraging $Q$-learning with a task-level state and reward. We implement our framework in simulation experiments and execute four challenging transfer tasks on robotic manipulation, demonstrating the efficacy of Diff-Transfer through comprehensive experiments. Supplementary and Videos are on the website ~\href{https://sites.google.com/view/difftransfer}{https://sites.google.com/view/difftransfer}

%% file: tex/intro.tex
\begin{figure}[t!]
 \centering
 \includegraphics[width=1.0\linewidth]{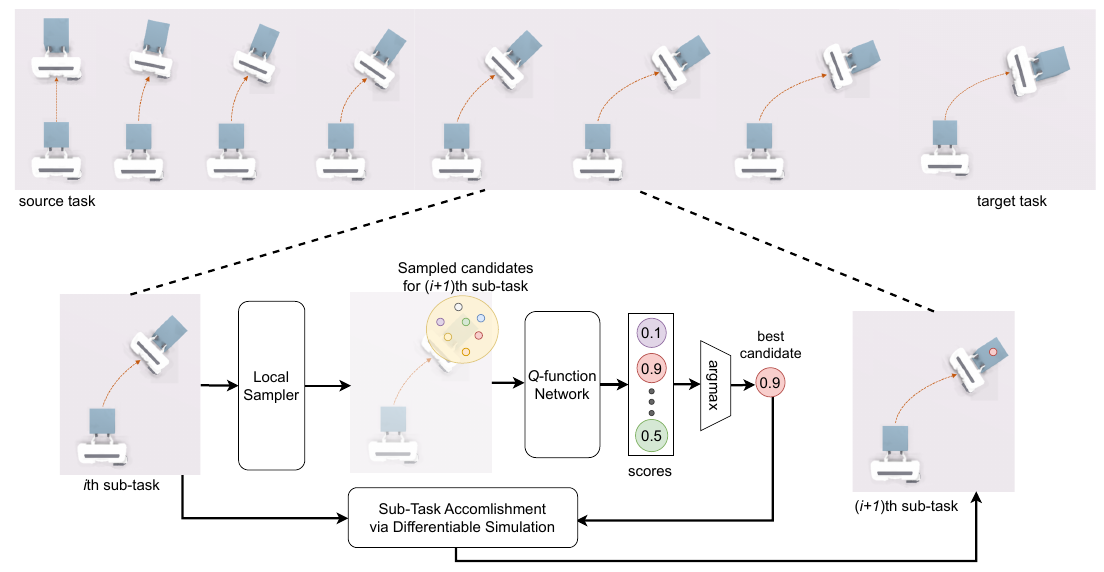}
  \caption{The overall approach of \textit{Diff-Transfer} includes a path of $L-1$ sub-tasks. \textit{Diff-Transfer} leverages \textit{Local Sampler}, \textit{$Q$-function Network} and \textit{argmax} function to select the best candidate to generate the $(i+1)$th sub-task given the $i$th sub-task, and learn the action sequence via differentiable physics simulation. }
\label{fig:teaser}
\end{figure}

The capacity for rapidly acquiring new skills in object manipulation is crucial for intelligent robots operating in real-world environments. One might wonder, how can robots efficiently learn manipulation skills across diverse objects? A straightforward approach would involve teaching a robot a new manipulation skill for every distinct object and task. However, this method lacks efficiency and is infeasible due to the vast variety of objects and possible robot interactions. Nonetheless, we could also notice that different manipulation skills may share common properties. As shown in Fig.~\ref{fig:teaser}, the one-directional pushing skill could be correlated to an object reorientation skill. Thus, it may be feasible to leverage prior knowledge acquired from one task to aid in learning another similar task. Transferring this prior knowledge and acquired skill set to new tasks could greatly enhance learning efficiency compared to starting from scratch.

Our intuition to solve this transfer learning problem is that Newton's Laws apply universally in our physical world. Therefore, when involved in similar tasks where objects are moved by similar poses, robots should interact with objects in similar ways.
In this way, efficiently leveraging the local information hidden in the variation of manipulation tasks could be the key to efficient task transfer learning.

In this paper, we investigate the problem of transferring manipulation skills between two object manipulation tasks.
Our proposed framework is depicted in Fig.~\ref{fig:teaser}. We approach this problem by interpolating the source task and target task by producing a large number of intermediate sub-tasks between them which gradually transform from the source task toward the target task.
These continuously and gradually transforming intermediate sub-tasks act as the bridge for transferring the action sequence from the source task to the target task. 

To better leverage the physical property associated with the object shape and pose transformation, we leverage differentiable simulation to capture model-based gradient information and use it in transforming robot action sequences. 
We introduce a refined $Q$-learning method for path planning in the pose transfer problem, where we use a high-level state and a well-designed reward to generate the path of seamlessly connected sub-tasks with a sample-based searching method.

We execute a series of challenging manipulation tasks using Jade\citep{gang2023jade}, a differentiable physics simulator designed for articulated rigid bodies. We undertake four tasks: \textit{Close Grill}, \textit{Change Clock}, \textit{Open Door}, and \textit{Open Drawer}. The outcomes demonstrate that our system surpasses prevalent baselines for transfer learning and direct transfer without path planning through differentiable simulation, highlighting the efficacy and merits of our approach. Additionally, we perform several ablation studies.

% We propose a virtual contact point search algorithm reformulating shape transfer into contact point transfer with certain searching goals, which avoids intricate mesh deformation problems. In this algorithm, we approach the problem with a gradient-based searching method leveraging the contact-based gradient feature in the differentiable physics simulation.
% Transfer learning studies transferring knowledge between different tasks assuming some knowledge gained from other tasks/resources is beneficial for learning a target task. In robotics, it has been mainly focused to improving reinforcement learning for robotic skills.

% While transfer learning has been extensively studied in the supervised learning domain, it is still active area for RL. 

In summary, we make the following contributions:
\begin{itemize}
\item We propose a systematic framework for model-based transfer learning, leveraging the differentiable physics-based simulation and applying our framework for pose transfer and object shape transfer.
\item We propose a novel path planning method for generating multiple sub-tasks in the task space and learning an action sequence for a new sub-task with the proximity property and leveraging $Q$-learning and differentiable physics simulation.
\item We conduct comprehensive experiments to demonstrate the effectiveness of our proposed transfer learning framework.
\end{itemize}

%% file: tex/relatedwork.tex
% We review related literature on key components in our approach, including model-based reinforcement learning, next best view, integration of differentiable physics-based simulation and rendering, and robotic manipulation. We describe how we are different from previous work.

\subsection{Differentiable simulation for manipulation. }
Significant advancements have been achieved in the field of differentiable physics engines, thanks to the evolution of automatic differentiation techniques~\citep{pytorch,team2016theano,hu2019taichi,bellCppAD,jax2018github,ceres-solver}. Various differentiable physics simulations have been developed for specific applications, such as rigid bodies~\citep{NEURIPS2018_842424a1,degrave2019differentiable,gang2023jade}, soft bodies~\citep{hu2019taichi,hu2019chainqueen,jatavallabhula2021gradsim,geilinger2020add,du2021_diffpd}, cloth~\citep{NEURIPS2019_28f0b864,qiao2020scalable,li2022diffcloth,yu2023diffclothai}, articulated bodies~\citep{werling2021fast,58d18a9e94704e8bb63f5e2959de61d6,qiao2021efficient}, and fluids~\citep{um2020solver,wandel2020learning,holl2020learning,Takahashi_Liang_Qiao_Lin_2021}.
Several studies have applied differentiable physics simulations to robotic manipulations. \citet{turpin2022grasp} focused on multi-fingered grasp synthesis, while \citet{lv2022sagci} guided robots in manipulating articulated objects. \citet{zhu2023difflfd, 10160649} enabled model-based learning from demonstrations by optimizing over dynamics, and \citet{lin2022diffskill,lin2022planning} targeted deformable object manipulation. \citet{gang2023jade} developed a differentiable simulation called \emph{Jade} for articulated rigid bodies with Intersection-Free Frictional Contact. 
%We adopt the

However, the incorporation of contact dynamics often results in non-convex optimization challenges due to discontinuities from contact mode switching~\citep{pmlr-v162-suh22b,antonova2022rethinking,zhu2023difflfd}. To mitigate this, contact-centric trajectory planning has been proposed~\citep{Mordatch2012Contact,7442110, 9560766, Gabiccini2018,zhu2023difflfd, jeannette_trajtree, 2023arXiv231003023H}, which plans both contact points and forces and generate manipulation actions afterward. Additionally, \citet{russ} introduced smoothing techniques for contact gradients and employed a convex quasi-dynamics model for feasible action searching. In alignment with existing research, our study utilizes differentiable physics simulations for the purpose of transferring robotic manipulation skills across different task spaces. % thereby facilitating model-based transfer learning.

\subsection{Transfer Learning in Robotics. }
Transfer learning has become a cornerstone in robotics, aiming to generalize skills across varying tasks, environments, or robotic platforms. Although still an open challenge, the majority of research has employed reinforcement learning (RL) for skill transfer~\citep{taylor2009transfer}.
Several approaches have been proposed to address this challenge. \citet{lazaric2008transfer, xu2021cocoi, jian2021adversarial,zhang2022learning,zhang2023efficient} utilize domain randomization during training to enhance agent robustness across diverse physical environments and to focus on task-relevant features. \citet{NEURIPS2018_9023effe, hu2023reboot} fine-tune reward and value functions on new tasks, while \citet{konidaris2007building}, \citet{liu2021learning}, \citet{zhang2023learning}, and \citet{9812312} directly adapt policies to new environments. \citet{finn2017model} introduces a meta-learning framework to improve agent adaptability across various tasks. \citet{chi2022irp} employs an iterative policy and approximates residual dynamics for runtime adaptation.
\citet{liu2022revolver, liu2022herd} use continuous robot interpolation and sequentially fine-tune RL policy to transfer skills from one robot to another.
Distinct from these approaches, our work adopts a model-based perspective for policy transfer. We utilize differentiable simulations to approximate physical dynamics and directly optimize pre-existing policies. We address the key differences between source and target environments as rewards where we accommodate varying manipulation goals that yield different reward functions. 
% \subsection{Shape Transfer}

%% file: tex/prodef.tex
% Given the policy $\pi_s^*$ which accomplishes a source task and a new task, we aim to learn an optimal policy $\pi_t^*$ for the target task. To be specific, for the source task, we have the initial trajectory $\{(s_r^{(i)}, s_o^{(i) })\}_{i=1}^T$ , $\{{a_r^{(i)}}\}_{i=1}^{T-1}$ where $s_r$ is the state of the robot, $s_o$ is the state of the object and $a_r$ is the action of the robot. We aim to learn a new action sequence $\{{a_r^{(i)\prime}}\}_{i=1}^{T-1}$ which accomplishes the target task.

% \yuqi{we should use the term "action/control" instead of "policy"} 

We consider two object manipulation tasks on a robot with $m$ joints.
We assume the source manipulation task is specified by the goal of object pose change $\Delta s_{\text{source}} \in \mathbb{R}^{6}$.
Suppose applying a given expert action sequence $ A_{\text{source}} = [a_{\text{source}}^{(t)}]_{t=1}^T$ on the task would yield a state-action trajectory $\tau_{\text{source}} = [(s_{r,{\text{source}}}^{(t)}, s_{o,{\text{source}}}^{(t)}, a_{\text{source}}^{(t)})]_{t=1}^T$ where $s_{r,{\text{source}}}^{(t)} \in \mathbb{R}^m$, $s_{o,{\text{source}}}^{(t)} \in \mathbb{R}^6 $, $a_{\text{source}}^{(t)} \in \mathbb{R}^m$ denotes robot state, object state and robot action at time $t$. 
We assume action sequence $A_{\text{source}}$ can successfully complete the task, i.e. moving the object from the starting pose $s_{o,{\text{source}}}^{(1)}$ to the goal pose $s_{o,{\text{source}}}^{(T)}= s_{o,{\text{source}}}^{(1)} + \Delta s_{\text{source}}$.
Our objective is to derive an action sequence $A_{\text{target}}=[a_{\text{target}}^{(t)}]_{t=1}^{T}$ that can successfully complete a new target manipulation task $\Delta s_{\text{target}}$ specified by the goal of object pose change $\Delta s_{\text{target}}$.

%% file: tex/approach.tex
\begin{comment}
\begin{figure}[t!]
 \centering
 \includegraphics[width=1.0\linewidth]{imgs/contact_point.pdf}
  \caption{\textbf{The change of contact points during shape transfer.}  
  The red and blue circles represent the two contact points for the two robot fingers respectively.
  % The positions of contact points changes during shape transfer.
  While the contact points are constrained to be on the object surface for source and target objects, we relax this constraint and allow the contact points to be not on the surface for intermediate objects.
  Therefore, we name it ``virtual contact point''.
  }
\label{fig:contact:point}
\end{figure}

\end{comment}
% \yuqi{we should use the term "action/control" instead of "policy"} 

We approach this problem by defining a path consisting of $L$ tasks % specified by the pose change goals
\begin{equation}
    \mathcal{P} = [ \Delta s_1,  \Delta s_2, \ldots, \Delta s_L]
\end{equation}
that connects the source and target tasks where $\Delta s_1 = \Delta s_{\text{source}}$ is the source task and $\Delta s_L = \Delta s_{\text{target}}$ is the target task.
Our approach consists of $L-1$ steps of action transfer.
At step $i$, our goal is to transfer a well-optimized action sequence $A_i$ on task $\Delta s_i$ to be a well-optimized action sequence $A_{i+1}$ on  the next task in the sequence $ \Delta s_{i+1}$.
For any $i$, we assume the difference between tasks $\Delta s_i$ and $ \Delta s_{i+1}$ is sufficiently small  so that the it is relatively easy to use local information such as differentiable simulation gradient to optimization for actions transfer.

\begin{equation} \label{eq:prop1}
    % ||s_{o,i}^{(T)} - s_{o,i+1}^{(T)}|| < \varepsilon_{1}
    ||\Delta s_i - \Delta s_{i+1}|| < \varepsilon_{1}
\end{equation}

where $\varepsilon_1$ denotes the upper limit between the final object state for two consecutive sub-tasks. This property is crucial to our gradient-based method in the following sub-section.

\subsection{How to accomplish a sub-task} \label{pose_sub-task}

% We can use a gradient-based method to solve the actions. Suppose the next sub-task goal pose is different from the current goal pose, then the sub-task actions are not too far away from the source actions, which suggests the idea of gradient descent. 

% Specifically, for each task, we can define a loss function $\mathcal{L}_{task}$, compute the gradient $\dfrac{\displaystyle \partial \mathcal{L}_{task}}{\displaystyle \partial a_r^{\prime}}$ leveraging the differentiable simulation Jade\cite{gang2023jade}, and then update actions.

% Within the realm of pose transfer, a sub-task is intrinsically defined by an altered objective pose of the object, represented as \(s_{o, \text{sub}}^{(T)}\).
% With one pair of sub-tasks $s_{o, i}^{(T)}$, $s_{o,i+1}^{(T)}$ on the pose path $\mathcal{P}_{pose}$, we will discuss how to learn a new optimal action sequence $\pi_{i+1}$ given $\pi_i$ and $\tau_i$, where $\tau_i$ is rollout trajectory $\{(s_{r,i}^{(t)}, s_{o,i}^{(t)}, a_i^{(t)} )\}_{t=1}^T$, using the optimal actions $\pi_i$ for $i$th task. %For clarity, $s_{o, i}^{(T)}$, $\pi_i$ are denoted as $s_{o, \text{source}}$

Our approach to deduce the requisite actions is through a gradient-based methodology. Under the assumption that the subsequent sub-task goal pose deviates from the current goal pose with a limited distance as described in Eq. \ref{eq:prop1}, we posit that the actions for the sub-task are in close proximity to the actions of the source. This postulation naturally lends itself to the application of gradient descent for optimization. We aim to optimize our current action sequence $\{ a_{\text{cur}}^{(t)}\}_{t=1}^T$, denoted as $A_{\text{cur}}$, with its initialization of $A_i$. The rollout trajectory based on $A_{\text{cur}}$ is denoted $\tau_{\text{cur}} = \{(s_{r, \text{cur}}^{(t)}, s_{o,\text{cur}}^{(t)}, a_{\text{cur}}^{(t)})\}_{t=1}^T$

To elaborate, for each specific task, we introduce a loss function, $\mathcal{L}_{task}$. 

\begin{equation} \label{eq:l_task}
    % \mathcal{L}_{task}\ = ||s_{o,\text{cur}}^{(T)} - s_{o,i+1}^{(T)}||^2
    \mathcal{L}_{task}\ = ||\Delta s_{\text{cur}} - \Delta s_{i+1}||^2
\end{equation}

where $\Delta s_{\text{target}}$ is the object pose change of $(i+1)$th sub-task goal and $\Delta s_{\text{cur}}$ is the object pose change of our rollout trajectory. We regard the task as accomplished if $\mathcal{L}_{task}$ is smaller than a certain threshold $\varepsilon_{t}$.

Utilizing the capabilities of the differentiable simulation framework Jade, we compute the gradient $\bigg\{\dfrac{ \partial \mathcal{L}_{task}}{ \partial a_{\text{cur}}^{(t)}} \bigg\}_{t=1}^T$, denoted as $\dfrac{\displaystyle \partial \mathcal{L}_{task}}{\displaystyle \partial A_{\text{cur}}}$. Subsequently, the current actions $A_{\text{cur}}$ are updated to minimize the task loss $\mathcal{L}_{task}$.

\begin{equation} \label{eq:pose_pi_update}
    A_{\text{cur}} \leftarrow A_{\text{cur}} - \eta \dfrac{\displaystyle \partial \mathcal{L}_{task}}{\displaystyle \partial A_{\text{cur}}}
\end{equation}

Thus we introduce Algorithm \ref{subgoal} as a function \textsc{transferStep}, since we will reuse this function in Section \ref{pose_sub-task}. It takes the trajectory $\tau_i$ for $i$th sub-task and the object pose change $\Delta s_{i+1}$ for $(i+1)$th sub-task as input. And it will output the optimized task loss $\mathcal{L}_{task}$, the boolean value $X$ indicating if the sub-task is successfully completed, and the rollout trajectory $\tau_{i+1}$ based on the optimized actions $A_{\text{cur}}$. If $X$ is True, then $A_{\text{cur}}$ is the desired $A_{i+1}$.
This algorithm iteratively refines the action sequence $A_{\text{cur}}$ over a maximum of $n_{epoch}$ iterations or until a convergence criterion is met.

\begin{algorithm}
  % \begin{tabularx}{1.0\linewidth}{c|c}
  %   \hline
  %   \textbf{Symbol} & \textbf{Description} \\
  %   \hline
  %   $\tau_s$ & source trajectory \\
  %   \hline
  %   $\tau_{\text{sub}}$ & sub-task trajectory \\
  %   \hline
  %   $s_r$/$s_r^{\prime}$  & source/sub-task state of robot  \\
  %   \hline
  %   $s_o$/$s_o^{\prime}$ & source/sub-task state of object \\
  %   \hline
  %   $a_r/a_r^{\prime}$ & source/sub-task action of robot \\
  %   \hline
  %   $s_{o, \text{sub}}^{(T)}$ & new sub-task goal \\
  %   \hline
  %   $\mathcal{L}_{task}$ & loss of task completion \\
  %   \hline
  %   $\eta$ & learning rate \\
  %   \hline
  %   $X$ & task success indicator \\
  %   \hline
  %   $\varepsilon_{pose}$ & loss threshold of pose transfer \\
  %   \hline
  %   \end{tabularx}
  \caption{Sub-Task Accomplishment}\label{subgoal}
  
  \begin{algorithmic}[1]
    \State \textbf{Input:} $\tau_i = \{(s_{r,i}^{(t)}, s_{o,i}^{(t)}, a_i^{(t)} )\}_{t=1}^T$, $\Delta s_{i+1}$
    \State \textbf{Output:} $\mathcal{L}_{task}$, $X$, $\tau_{{i+1}}$

    \Function{transferStep}{$\tau_s, \Delta s_{i+1}$}
    \State $s_{r, \text{cur}}^{(1)} \leftarrow s_{r,i}^{(1)},a_{\text{cur}}^{(t)} \leftarrow a_i^{(t)}, t=1,2,\ldots,T$ %\textcolor{darkgray}{// Initialize}
        \For {$e$ \textbf{in} $1, 2, \dots, n_{epoch}$ }%\textcolor{darkgray}{// Training until convergence}

            \For{$t$ \textbf{in} $1, 2, \dots, T-1$} 
                \State $(s_{r,\text{cur}}^{(t+1)},s_{o, \text{cur}}^{(t+1)})\leftarrow \textbf{simulate}(s_{r,\text{cur}}^{(t)}, s_{o, \text{cur}}^{(t)}, a_{\text{cur}}^{(t)})$
            \EndFor  
            \State $\Delta s_{\text{cur}} \leftarrow s_{o, \text{cur}}^{(T)} - s_{o, \text{cur}}^{(1)}$
            \State $\mathcal{L}_{task} \leftarrow  ||\Delta s_{\text{cur}} - \Delta s_{i+1}||^2$
            % \State $a_r^{\prime} \leftarrow a_r^{\prime} - \eta \dfrac{\displaystyle \partial \mathcal{L}_{task}}{\displaystyle \partial a_r^{\prime}}$ \textcolor{darkgray}{ // Update the action of robot.}
            \State  $A_{\text{cur}} \leftarrow A_{\text{cur}} - \eta \dfrac{\displaystyle \partial \mathcal{L}_{task}}{\displaystyle \partial A_{\text{cur}}}$
            \If{$\mathcal{L}_{task} \leq \varepsilon_{t}$}
            \State \Return $\mathcal{L}_{task}$, \textbf{True}, $\{(s_{r,\text{cur}}^{(t)}, s_{o, \text{cur}}^{(t)}, a_{\text{cur}}^{(t)})\}_{t=1}^T$
            \EndIf
        \EndFor
    \State \Return $\mathcal{L}_{task}$, \textbf{False}, $\{(s_{r,\text{cur}}^{(t)}, s_{o, \text{cur}}^{(t)}, a_{\text{cur}}^{(t)})\}_{t=1}^T$
    \EndFunction
  \end{algorithmic}
\end{algorithm}

\subsection{Sub-Tasks Generation} \label{pose_path}

% \yuqi{Shall we describe the network as Q-learning?} \\
% \yuqi{Input: sub-task goal end state \\ output: loss/reward for this end state}

% The approach described by Algorithm \ref{subgoal} is predominantly efficient when the target goal pose is proximate to the source goal pose. For scenarios where there's a significant divergence between the final target and source goal poses, the optimization process could become arduous, necessitating excessive iterations or potentially resulting in non-convergence. To circumvent this, we transpose our problem into one of path searching.

% Given the source goal $s_{o}^{(T)}$ and the final target $s_{o, \text{target}}^{(T)}$, our objective is to delineate a path $(s_{o}^{(T)}, s_{o, 1}^{(T)}, s_{o, 2}^{(T)}, \dots, s_{o, l-1}^{(T)}, s_{o, \text{target}}^{(T)})$, ensuring that the distance between any two subsequent sub-task poses remains small enough as well as guaranteeing transferability.

Given Algorithm \ref{subgoal} and the path $\mathcal{P}$, it is easy to compute the optimized actions $A_t$ for our target task, since we can use dynamic programming to optimize $A_{i+1}$ based on $A_i$. The only problem is to generate one feasible path $\mathcal{P}$ where not only the property in Eq. \ref{eq:prop1} holds but also the Algorithm \ref{subgoal} tends to return the successful result with optimized action sequence $A_{i+1}$ and the corresponding trajectory $\tau_{i+1}$ for $(i+1)$th sub-task for each index $i$. This reduces the problem into a path planning problem in the goal pose space where each node in the space denotes a goal final object state and we aim to build a path connecting the source goal pose and the target one.

While there are lots of traditional path-planning algorithms in 3-D Euclidean space, they fail to solve our problem because the goal pose space is in a higher dimension and the obstacle is harder to detect. We introduce our innovative reinforcement learning method by predicting the difficulty of sub-tasks using a refined $Q$-function neural network $Q(x;\theta)$ parameterized by $\theta$. Instead of taking input of the conventional state and action at time $t$, the network takes a high-level state input $x$, which could be any object pose change like $\Delta s_{\text{target}}$. The output $r$ would be the estimated reward.

Unlike traditional RL problems with clear task rewards, the reward in our problem needs an elaborate design because we are performing path planning on a higher task-space level. We introduce the reward function as 

\begin{equation}
    \label{eq:reward}
    r(x) = -(\lambda_t \cdot \mathcal{L}_{task} + \lambda_{d} \cdot ||x - \Delta s_{\text{target}}||^2)
\end{equation}

To illustrate this equation, the first term $\mathcal{L}_{task}$ is computed using Eq. \ref{eq:l_task} where $\Delta s_{i+1}$ is given as $x$ and $\Delta s_{\text{cur}}$ is given by the optimized actions $A_{\text{cur}}$ for sub-task goal $x$. The second term $||x - \Delta s_{\text{target}}||^2$, shortly as $\mathcal{L}_{dis}$, describes the distance from pose change $x$ to the target pose change $\Delta s_{\text{target}}$. Finally, $\lambda_t$ and $\lambda_{d}$ are weight coefficients to balance these two terms. Therefore, such reward results in a better path-planning algorithm because when the reward is high, both the task loss $\mathcal{L}_{task}$ and the distance to target goal $\mathcal{L}_{dis}$ are low.

Suppose we have the accurate $Q(x;\theta)$ network, we can generate the path $\mathcal{P}$ in either a gradient-based way or a sample-based way. We employ the sampled-based approach for the current pose transfer problem to increase the robustness of stochastic noise from the inaccurate network in reality. In detail, given $i$th sub-task with a pose change $\Delta s_{i}$, we sample $n$ vectors $\{x_j\}_{j=1}^n$, denoted as $S$, in the task space in the neighbourhood of the $i$th sub-task goal $\Delta s_{i}$, so that

\begin{equation}
    ||\Delta s_{i} - x_j|| < \varepsilon_{sample}, j = 1,2,\dots,n
\end{equation}

where $\varepsilon_{sample}$ is the radius of the neighbourhood. In these $n$ candidates for the $(i+1)$ sub-task, we choose the best one $k$ based on our current knowledge to maximize the reward $r_k$

\begin{equation}
    k = \arg\max_j r_j, j = 1,2,\dots,n
\end{equation}

Once we get the best candidate $x_k$, we call the function \textsc{transferStep} in Algorithm \ref{subgoal}, in an attempt to optimize an action sequence $A_{i+1}$ for the given $(i+1)$th sub-task. Should this process be successful, we shall continue to generate the next sub-task recursively until the target goal is attained. Otherwise, we shall discard this candidate $x_k$ and find an alternative best candidate from $S$ iteratively, as is shown in Algorithm \ref{path-search}.

To learn an approximate network $Q(x; \theta)$, we maintain a dataset $D$ dynamically during the path-planning process. Each time after we call the \textsc{transferStep} function and get more information about the task space, we add the data pair $(x_k, r_k)$ into $D$, update $\theta$ with the $Q$-learning method to gain a better network and proceed on path planning.

\subsection{Implementation Details} \label{sec:impl}

In this section, we discuss the implementation details of \textit{Diff-Transfer} in Algorithm \ref{path-search}. To begin with, we pre-train our network $Q(x;\theta)$ with a refined initial reward in Eq. \ref{eq:reward}, where $\mathcal{L}_{task}$ is set to a certain constant $c_t$ because we cannot know the difficulty of any sub-task beforehand. Specifically, we generate labels $(x_{\text{pre}}, r_{\text{pre}})$ randomly to build a dataset $D_{\text{pre}}$ and use it to fit $\theta$ using a supervised learning method via minimizing the loss $l_{\text{pre}}(\theta) = ||Q(x_{\text{pre}};\theta) - r_{\text{pre}}||^2$. With online dataset $D=\{(x_k, r_k)\}_{k=1}^{m}$ collected during execution of our path-planning method, network parameters $\theta$ will be fine-tuned  to minimize the loss $l(\theta)=||Q(x_k;\theta) - r_k||^2$. It is worth noting that $D$ doesn't contain data from $D_{\text{pre}}$ because data in $D$ collected from rollouts in simulation reflect the actual rewards of sub-tasks while $D_{\text{pre}}$ just provides a rough estimation under the hypothesis that all sub-tasks have same difficulties, which is hardly true in the real transfer problem.

\begin{algorithm}
  \caption{$Q$-function Network Guided Path Planning}\label{path-search}
  \begin{algorithmic}[1]
    % \State \textbf{Input:}target pose $s_o^{(t)}$ and subgoal pose $s_{o, \text{subgoal}^{(t)}}$ with the trajectory $\tau_s$.
    % \State Set a neural network $f_\theta$, taking a 6D vector (pose) as input and outputting a scalar value (estimated loss).
    % \State Pretrain the network with $f_\theta(P) = Th_{task} + \lambda \cdot ||P - P_t||^2 $.
    % \State $D = \varnothing$.
    % \State Run \Call{pathSearch}{$P_s, P_t, \tau_s$}.
    \Function{pathSearch}{$\tau_i, \Delta s_{\text{target}}$}
        \If {$||\Delta s_{i} - \Delta s_{\text{target}}|| \leq \varepsilon_{pose}$ }
        \State \textbf{return} $\tau_i$
        \EndIf
        \State Randomly sample $n$ vectors $S \leftarrow \{x_j\}^n_{j=1}$ in the neighbourhood of $\Delta s_{i}$
        \State $r_j \leftarrow Q_\theta(x_j), j = 1, 2... n$.
        \While{$S \neq \varnothing$}
        % \State Choose the smallest $y_k$
        \State $k \leftarrow \arg\max_j{r_j}$
        \State $\mathcal{L}_{task}$, $X$, $\tau_{i+1}$ $\leftarrow$  \Call{transferStep}{$\tau_i, x_k$}
        \State $\mathcal{L}_{dis} \leftarrow ||x_k - \Delta s_{\text{target}}||^2$
        \State $r_k \leftarrow -(\lambda_t \cdot \mathcal{L}_{task} + \lambda_{d} \cdot \mathcal{L}_{dis})$
        \State $D \leftarrow D \cup \{(x_k, r_k )\}$ 
        \State Update $\theta$ using dataset $D$
        \If {$X = \textbf{True}$}
        \State \Call{pathSearch}{$\tau_{i+1}, \Delta s_{\text{target}}$}
        \Else
        \State $S \leftarrow S - \{x_k\}$
        \State \textbf{continue}
        \EndIf
        \EndWhile
        \State \Return failure
    \EndFunction
    \end{algorithmic}
\end{algorithm}

%% file: tex/exp.tex
In this section, we present a rigorous experimental framework meticulously designed to elucidate the effectiveness of our proposed system \textit{Diff-Transfer}. This exhaustive evaluation encompasses an assessment of the system's performance across diverse conditions, while also subjecting it to rigorous scrutiny in the presence of unforeseen challenges. The tests conducted in this study are geared towards offering a comprehensive panorama of the system's capabilities. Our foremost objective is to substantiate the theoretical foundations expounded earlier and establish a seamless connection between theory and practical implementation, thereby affirming the system's scalability and adaptability across a multitude of application domains.

%In this section, we delve into a meticulous experimental framework designed to shed light on the efficacy of our proposed system. This comprehensive evaluation not only gauges the system's performance under varying conditions but also scrutinizes its robustness in the face of unforeseen challenges. Our tests aim to provide a holistic overview of the system's capabilities and our primary objective is to validate the theoretical constructs presented earlier and bridge the gap between theory and practical application, ensuring the system is both scalable and adaptable to a myriad of applications.

\subsection{Experimental Setup}
\subsubsection{Simulation Setting}

We choose multiple manipulation tasks from RLBench \citep{james2020rlbench} and adapt the environment to the Jade\citep{gang2023jade} simulation. Specifically, we acquire the trajectory of states for each task, along with the objects' Unified Robot Description Format (URDF) files and corresponding mesh files. Actions are computed utilizing inverse dynamics and optimization within Jade, providing us with a comprehensive initial trajectory of both states and actions, denoted as $\tau_{\text{source}}$. 

\subsubsection{Evaluation Metric} 

We employ the number of iterations $N$ in the optimization loop to evaluate the efficiency of our methods and compare the results. We also report the distance $d$, which is a task-related metric describing the completeness of manipulation. For each specific manipulation task, we run $5$ times our method to reduce the effect of randomness and report the mean value for both the iterative steps and the distance as $\bar{N}$ and $\bar{d}$, and the standard deviation as $\sigma_N$ and $\sigma_d$. 

\subsubsection{Manipulation Skill Transfer Tasks} 
\paragraph{\textit{Close Grill}}

The robot is required to close a grill lid. This task is considered successful if the grill lid has been rotated to close. The distance $d$ describes the distance from the final angle of the grill lid joint to the target angle, with a unit of degrees. 

% \paragraph{Close Drawer}

% The robot is required to close a drawer. The chest has 3 drawers. This task is considered successful if the specific drawer has been pushed into the chest.

\paragraph{\textit{Change Clock}}

The robot is required to change a clock. This task is considered successful if the clock pointer has been revoluted to a specific orientation. The distance $d$ describes the distance from the final angle of the clock pointer to the target angle, with a unit of degrees.  

\paragraph{\textit{Open Door}}

The robot is required to open a door. This task is considered successful if the door has been rotated to a specific orientation from the door frame. The distance $d$ describes the distance from the final angle of the door to the target angle, with a unit of degrees.

\paragraph{\textit{Open Drawer}}

The robot is required to open a drawer. The chest has 3 drawers. This task is considered successful if the specific drawer has been pulled out from the chest. The distance $d$ describes the distance from the final translation of the drawer to the target angle, with a unit of meters. 

% \subsubsection{Object Transfer Tasks}

% \paragraph{Pushing}

% The robot is required to push an object. The task is considered successful if the object reaches the target pose with the correct position and orientation. The source object is a cube and the target object is a pyramid.

% % \paragraph{Grasping}

% % The robot is required to grasp an object. The task is considered successful if the object is grasped and lifted. The source object is XXX and the target object is XXX.

% \paragraph{Rotating}

% The robot is required to rotate an object on the table. The task is considered successful if the object is rotated to the specified orientation. The source object is a cube and the target object is a pyramid.
\begin{comment}

\begin{figure}
    \centering
    \includegraphics[width= \linewidth]{imgs/pose_transfer2.png}
    \caption{Visualization of experiment setup for \textit{Close Grill}, \textit{Change Clock}, \textit{Open Door} and \textit{Open Drawer}.}
    \label{fig:pose_transfer}
\end{figure}
\end{comment}

\begin{figure}
\centering
    \subfloat[]{\includegraphics[width=0.24\linewidth]{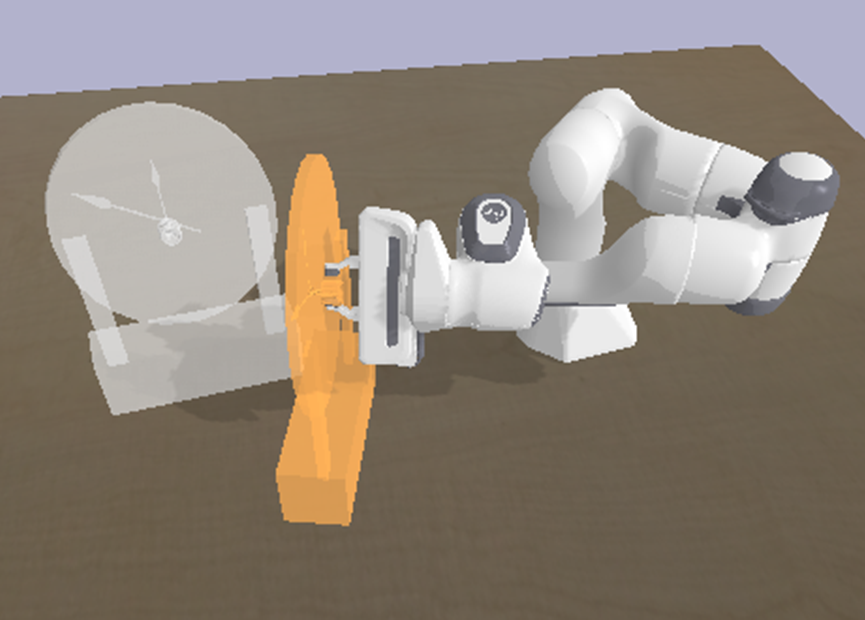}
    }
    \subfloat[]{\includegraphics[width=0.24\linewidth]{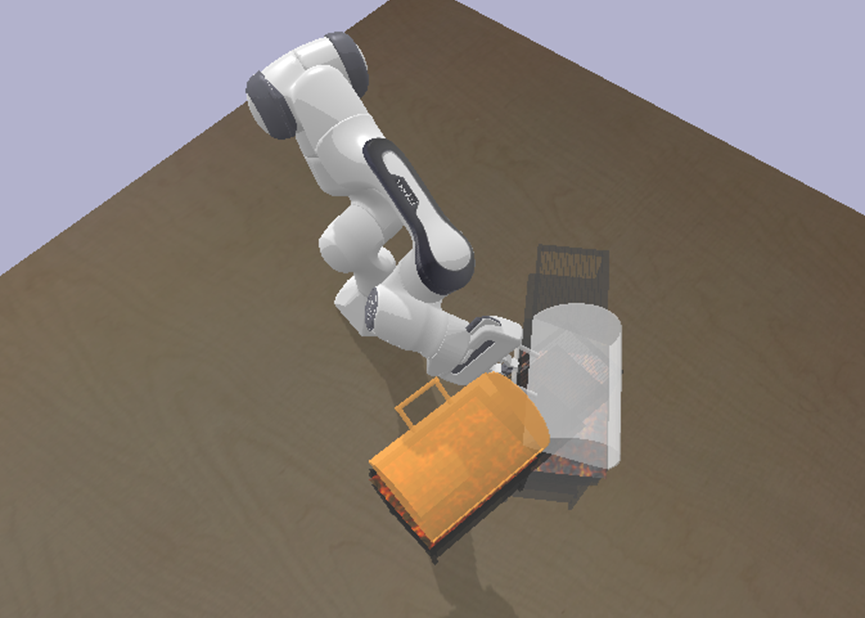}
    }    
    \subfloat[]{\includegraphics[width=0.24\linewidth]{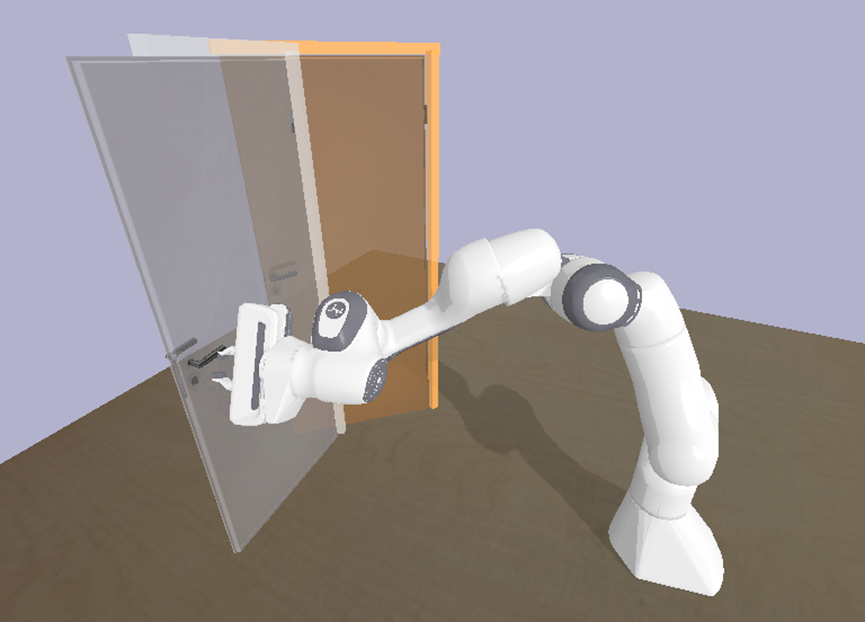}
    }    
    \subfloat[]{\includegraphics[width=0.24\linewidth]{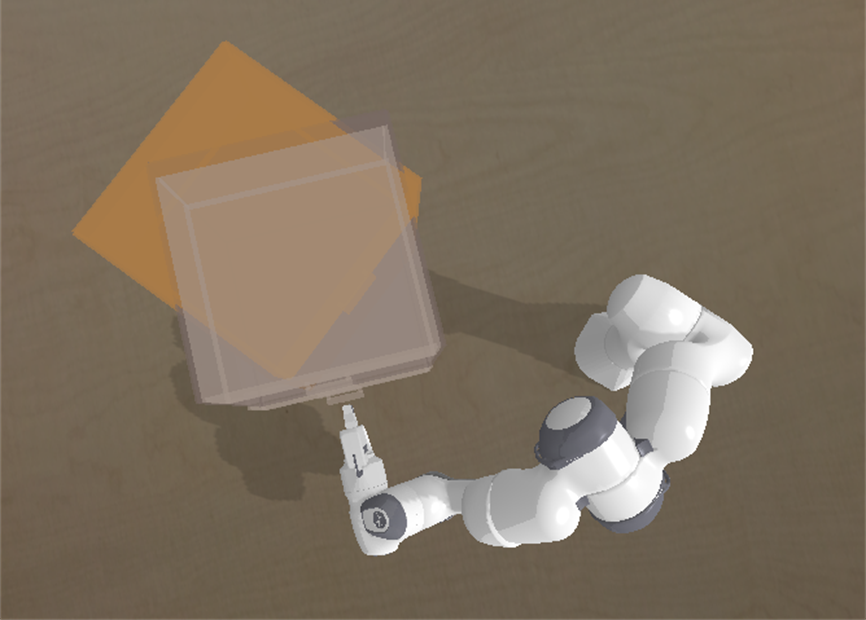}
    }    
    \caption{Source Task (grey object) and Target Task (orange object) for (a) \textit{Change Clock}, (b) \textit{Close Grill},  (c) \textit{Open Door}, and (d) \textit{Open Drawer}. }
    \label{fig:pose_transfer2}
\end{figure}

\subsubsection{Implementation Details}

To illustrate the details presented in Section \ref{sec:tech}, we define $\Delta s_i$, the objective of the $i$th sub-task, as the base pose change of the manipulated object from its pose in the source task. This definition slightly diverges from the description in Section \ref{probdef}, as these intricate manipulation tasks require the robot to manipulate the object’s joint, rather than altering its pose by pushing.

We employ a three-layer MLP to implement the Q-function network $Q(x;\theta)$. Rather than directly utilizing the reward function in Eq. \ref{eq:reward}, we characterize the output network as an estimated loss with a value of $-r(x)$, explaining why the landscapes in Fig. \ref{fig:landscape} exhibit a minimum area instead of a maximum, a point to be discussed in subsequent Section \ref{sec:exp_results}.

\begin{comment}
\begin{center}
\begin{table}[]
\centering
\resizebox{1\linewidth}{!}{%
\begin{tabular}{c|ccc|ccc|ccc}
\toprule
Method&
\multicolumn{3}{c}{Direct Transfer}&\multicolumn{3}{|c}{MAML}&\multicolumn{3}{|c}{Path Search}\\
\midrule
Task Name &Steps&Loss&Succ&Steps&Loss&Succ&Steps&Loss&Succ\\ 

  \midrule
Change Clock & 500 &$1.18*10^{-1}$&$\times$& 5000  & $2.41*10^{-2}$&$\times$& \textbf{8}   & $\mathbf{7.51*10^{-3}}$&\checkmark\\ 
\cdashline{1-10}[2pt/2pt]
\rule{0pt}{10pt}
Close Grill  & 500 & $2.22*10^{-2}$ &$\times$&4800 & $1.01*10^{-3}$&\checkmark& \textbf{92}  & $\mathbf{2.22*10^{-4}}$ &\checkmark\\ 
\cdashline{1-10}[2pt/2pt]
\rule{0pt}{10pt}
% Close Drawer & \textbf{1}        & 8.46E-05  & 5            & 7.93E-05   \\ \midrule
Open Door    & 255 & $5.95*10^{-4}$&\checkmark& 5000 &$3.04*10^{-1}$ &$\times$&\textbf{10}  & $\mathbf{4.21*10^{-4}}$ &\checkmark\\ 
\cdashline{1-10}[2pt/2pt]
\rule{0pt}{10pt}
Open Drawer  &500 & $3.78*10^{-2}$  &$\times$&4300 & $2.00*10^{-5}$&\checkmark&\textbf{203} & $\mathbf{9.55*10^{-3}}$ &\checkmark \\ 
\bottomrule
\end{tabular}%
}
\caption{Iterations and Final Task Loss of Direct Transfer and Path Search, TODO: update data}
\label{tab:pose_results}
\end{table}
\end{center}
\end{comment}

% \usepackage{booktabs}
% \usepackage{graphicx}
\begin{table}[]
\centering
\resizebox{\textwidth}{!}{%
\begin{tabular}{@{}ccccclcccccc@{}}
\toprule
Method &
  \multicolumn{4}{c}{\textit{Diff-Transfer (Ours)}} &
  \multicolumn{2}{c}{\textit{MAML}} &
  \multicolumn{2}{c}{\textit{DMG}} &
  \multicolumn{3}{c}{\textit{Direct Transfer}} \\ \midrule
Task Name &
  $\bar{N}$ &
  $\sigma_N$ &
  $\bar{d}$ &
  $\sigma_d$ &
  \multicolumn{1}{c}{$d$} &
  success &
  $d$ &
  success &
  $N$ &
  $d$ &
  success \\ \midrule
\multicolumn{1}{c|}{\textit{Change Clock}} &
  \textbf{55.6} &
  61.1 &
  \textbf{3.72} &
  \multicolumn{1}{c|}{1.38} &
  10.27 &
  \multicolumn{1}{c|}{$\times$} &
  27.46 &
  \multicolumn{1}{c|}{$\times$} &
  1000+ &
  19.66 &
  $\times$ \\
\multicolumn{1}{c|}{\textit{Close Grill}} &
  \textbf{66.4} &
  11.5 &
  \textbf{1.80} &
  \multicolumn{1}{c|}{0.55} &
  18.54 &
  \multicolumn{1}{c|}{$\times$} &
  56.71 &
  \multicolumn{1}{c|}{$\times$} &
  1000+ &
  8.53 &
  $\times$ \\
\multicolumn{1}{c|}{\textit{Open Door}} &
  \textbf{57.8} &
  38.2 &
  \textbf{0.64} &
  \multicolumn{1}{c|}{0.43} &
  9.20 &
  \multicolumn{1}{c|}{$\times$} &
  41.91 &
  \multicolumn{1}{c|}{$\times$} &
  255 &
  1.40 &
  $\checkmark$ \\
\multicolumn{1}{c|}{\textit{Open Drawer}} &
  \textbf{123.8} &
  103.9 &
  \textbf{0.06} &
  \multicolumn{1}{c|}{0.00} &
  0.08 &
  \multicolumn{1}{c|}{$\times$} &
  0.18 &
  \multicolumn{1}{c|}{$\times$} &
  1000+ &
  0.12 &
  $\times$ \\ \bottomrule
\end{tabular}%
}
\caption{Experiment Results for \textit{MAML}, \textit{DMG}, \textit{Direct Transfer} and our \textit{Diff-Transfer}. \textit{Diff-Transfer} is executed using 5 distinct random seeds.}
\label{tab:baseline}
\end{table}

% \usepackage{booktabs}
% \usepackage{graphicx}
\begin{table}[]
\centering
\resizebox{\textwidth}{!}{%
\begin{tabular}{@{}cccccccccccc@{}}
\toprule
Method &
  \multicolumn{4}{c}{\textit{Diff-Transfer}} &
  \multicolumn{4}{c}{\textit{Diff-Transfer ($\lambda_t=0$)}} &
  \multicolumn{3}{c}{\textit{Linear Interpolation}} \\ \midrule
Task Name &
  $\bar{N}$ &
  $\sigma_N$ &
  $\bar{d}$ &
  $\sigma_d$ &
  $\bar{N}$ &
  $\sigma_N$ &
  $\bar{d}$ &
  $\sigma_d$ &
  $N$ &
  success &
  $d$ \\ \midrule
\multicolumn{1}{c|}{Change Clock} &
  55.6 &
  61.1 &
  3.72 &
  \multicolumn{1}{c|}{1.38} &
  \textbf{51.0} &
  28.7 &
  \textbf{3.23} &
  \multicolumn{1}{c|}{1.70} &
  68.0 &
  \checkmark &
  5.43 \\
\multicolumn{1}{c|}{Close grill} &
  \textbf{66.4} &
  11.5 &
  \textbf{1.80} &
  \multicolumn{1}{c|}{0.55} &
  96.6 &
  28.4 &
  2.45 &
  \multicolumn{1}{c|}{0.55} &
  157.0 &
  \checkmark &
  3.36 \\
\multicolumn{1}{c|}{Open Door} &
  \textbf{57.8} &
  38.2 &
  \textbf{0.64} &
  \multicolumn{1}{c|}{0.43} &
  185.4 &
  118.3 &
  2.78 &
  \multicolumn{1}{c|}{2.16} &
  113.0 &
  \checkmark &
  4.11 \\
\multicolumn{1}{c|}{Open Drawer} &
  \textbf{123.8} &
  103.9 &
  \textbf{0.06} &
  \multicolumn{1}{c|}{0.00} &
  527.0 &
  712.0 &
  \textbf{0.06} &
  \multicolumn{1}{c|}{0.00} &
  309.0 &
  $\times$ &
  0.38 \\ \bottomrule
\end{tabular}%
}
\caption{Experiment Results for \textit{Diff-Transfer (Ours)}, \textit{Diff-Transfer ($\lambda_t=0$)}, and \textit{Linear Interpolation}. Both \textit{Diff-Transfer} and \textit{Diff-Transfer ($\lambda_t=0$)} are executed using 5 distinct random seeds.}
\label{tab:ablation}
\end{table}

% \begin{figure}[]
%     \centering
%     \subfigure[Close Grill Task]{\includegraphics[width=.4\linewidth]{imgs/close_grill_view.png}}
%     \hfill
%     \subfigure[Network Landscape for Close Grill]{\includegraphics[width=.4\linewidth]{imgs/close_grill_landscape.png}}
%     \\
%     % \subfigure[Close Drawer Task]{\includegraphics[width=.4\linewidth]{imgs/close_drawer_view.png}}
%     % \hfill
%     % \subfigure[Network Landscape for Close Drawer]{\includegraphics[width=.4\linewidth]{imgs/close_drawer_landscape.png}}
%     % \\
%     \subfigure[Change Clock Task]{\includegraphics[width=.4\linewidth]{imgs/change_clock_view1.png}}
%     \hfill
%     \subfigure[Network Landscape for Change Clock]{\includegraphics[width=.4\linewidth]{imgs/change_clock_landscape.png}}
%     \\
%     \subfigure[Open Door Task]{\includegraphics[width=.4\linewidth]{imgs/open_door_view.png}}
%     \hfill
%     \subfigure[Network Landscape for Open Door]{\includegraphics[width=.4\linewidth]{imgs/open_door_landscape.png}}
%     \\
%     \subfigure[Open Drawer Task]{\includegraphics[width=.4\linewidth]{imgs/open_drawer_view.png}}
%     \hfill
%     \subfigure[Network Landscape for Open Drawer]{\includegraphics[width=.4\linewidth]{imgs/open_door_landscape.png}}

%     \caption{Experiments on tasks for Goal Pose Transfer}
%     \label{fig:pose_transfer}
% \end{figure}

\subsection{Baseline} 

\paragraph{DMP}
DMP (Dynamic Movement Primitives) is a method for learning and reproducing complex dynamic movement skills in robots and other systems, making it easier for them to perform tasks such as reaching and grasping objects. Specifically, for a transfer task, we use the robot trajectory of the source task to fit the dmp function, modify the object target on the target task and reproduce the motion trajectory.

\paragraph{MAML}
Model-agnostic meta-learning (MAML) is a meta-learning algorithm that enables machine learning models to quickly adapt to new tasks with minimal training data by learning good initializations that can be fine-tuned for specific tasks, making it highly applicable to a variety of applications. application. Specifically, for a transfer task, we perform learning on $4$ source tasks and perform trajectory prediction on a target task. In our experiments, the trained policy is a two-layer MLP network with $128$ hidden units in each layer. We use the adam optimizer and SGD loss function to train the policy for $1000$ epochs. In each epoch, we perform task-level training and meta-training. During each task-level training, we sample $20$ trajectories on four source tasks to update the parameters of the task-level strategy. During each meta-training, we use task-level update parameters to sample $5$ trajectories on $4$ source tasks and update the policy parameters. We will train the final trained policy on the target task for $20$ epochs to fine-tune the parameters, and calculate whether the policy given at this time can complete the target task.

\paragraph{Direct Transfer}

To demonstrate the efficacy of our path-searching method, we assess the direct transferring technique on each task, using it as one of the baselines, denoted as \textit{Direct Transfer}. Contrary to constructing a path where the source task and the target task are cohesively linked via several intermediate sub-tasks as in Algorithm \ref{path-search}, \textit{Direct Transfer} solely endeavors to optimize an action sequence for the target task, directly drawing from the source task trajectory through differentiable simulation, as outlined in Algorithm \ref{subgoal}.

\subsection{Experiment Results} \label{sec:exp_results}

\begin{figure}
\centering
    \subfloat[]{\includegraphics[width=0.24\linewidth]{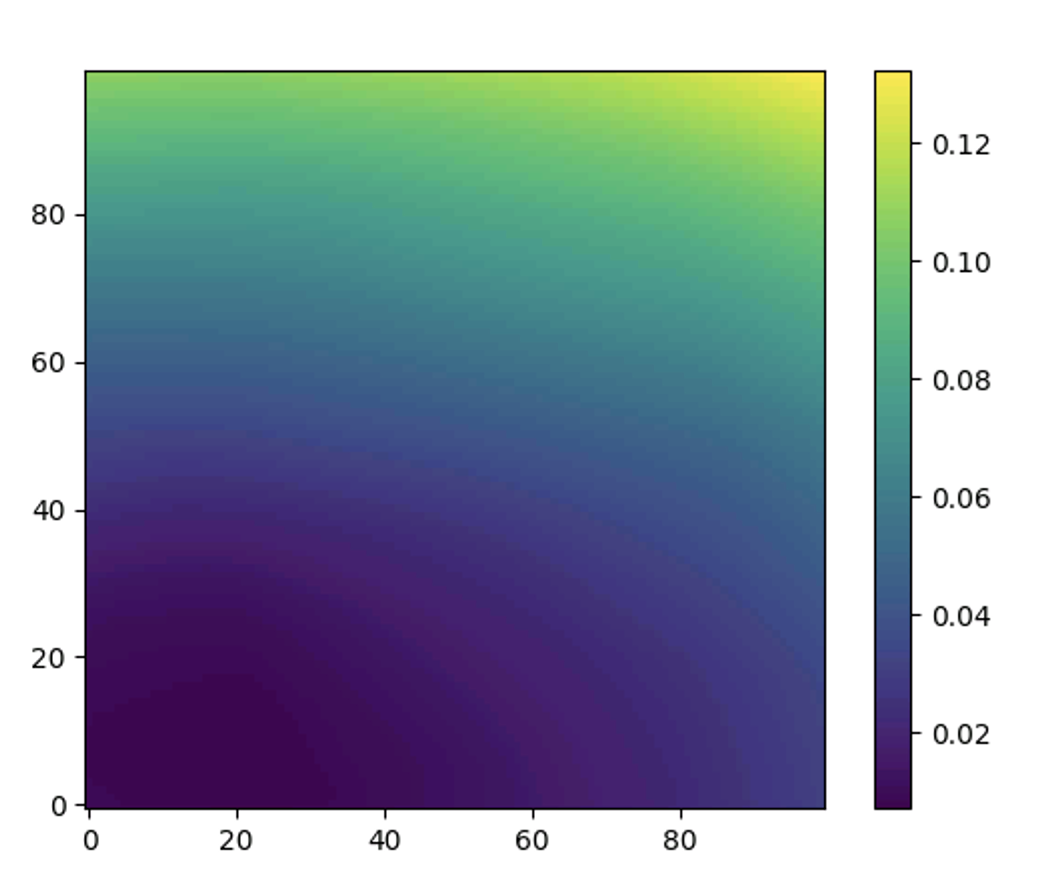}
    }
    \subfloat[]{\includegraphics[width=0.24\linewidth]{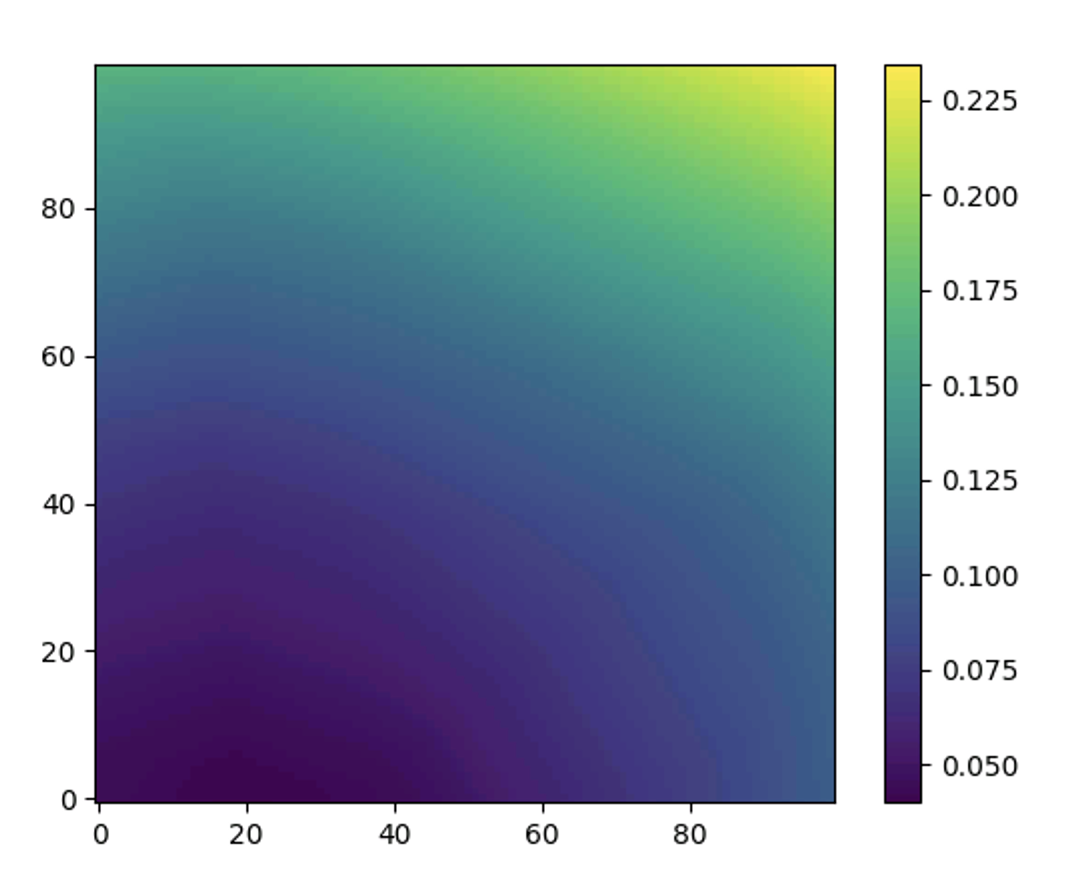}
    }    
    \subfloat[]{\includegraphics[width=0.24\linewidth]{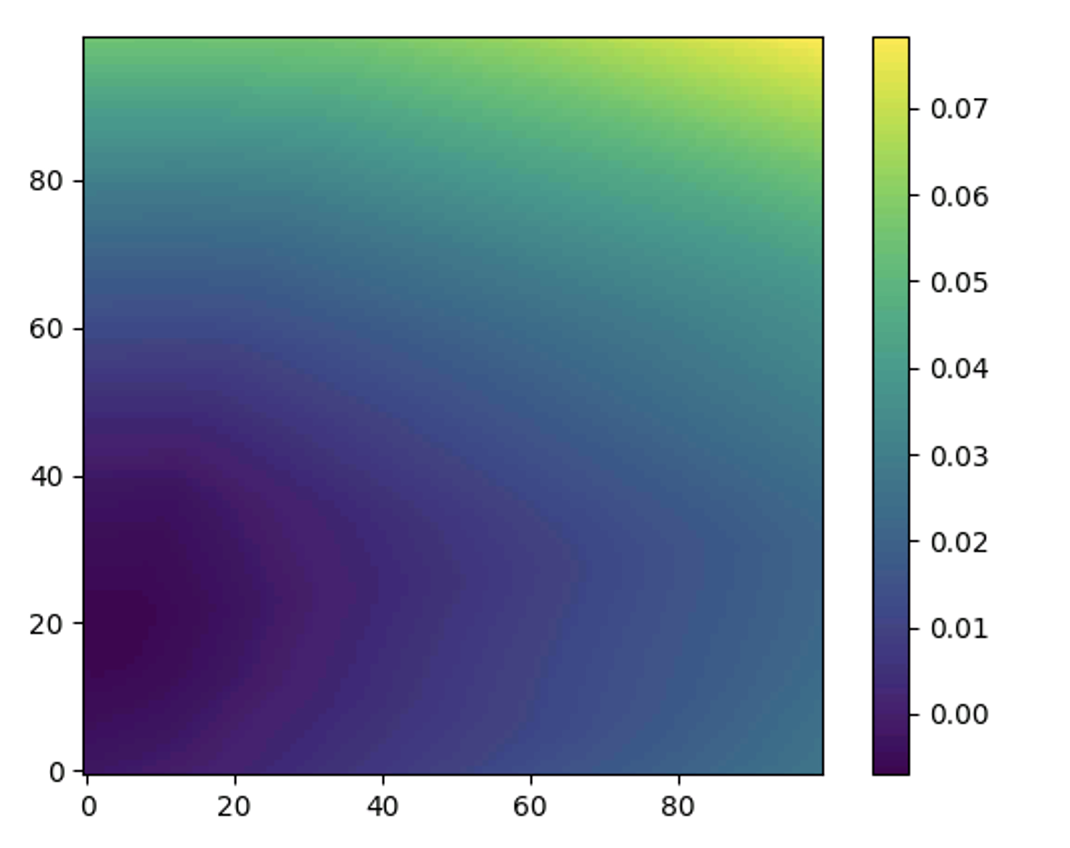}
    }    
    \subfloat[]{\includegraphics[width=0.24\linewidth]{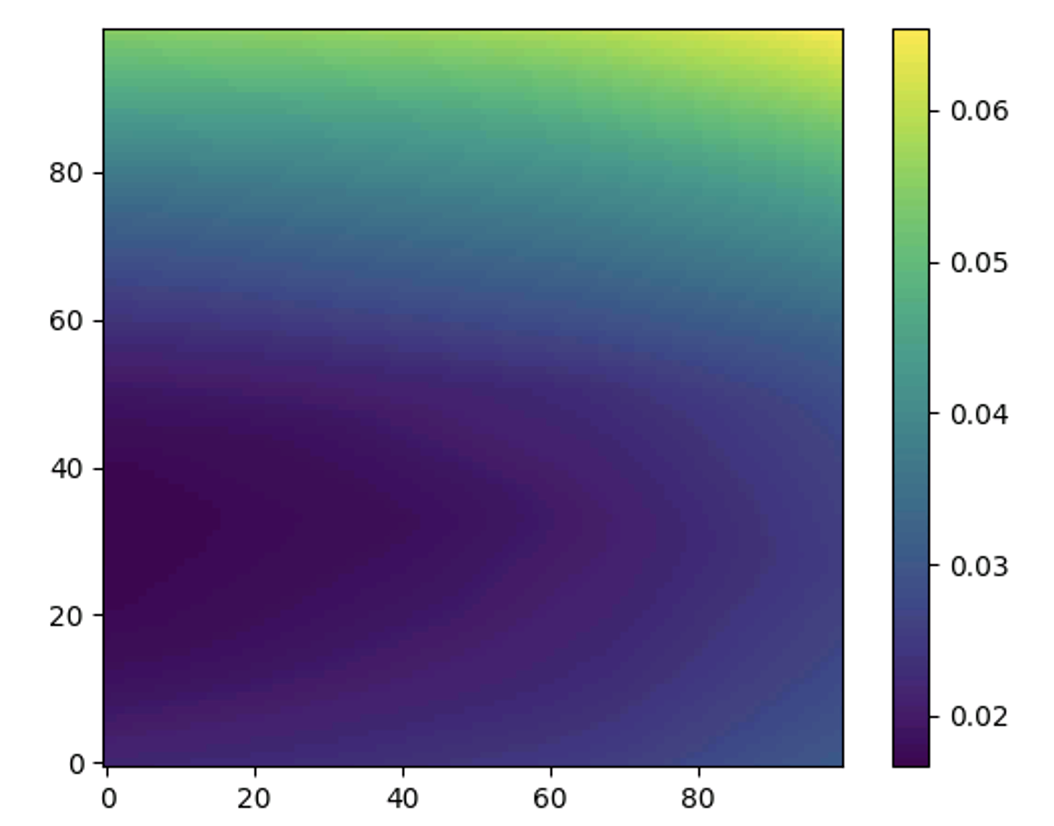}
    }    
    \caption{Visualization of learned $Q$-function Landscapes for (a) \textit{Change Clock}, (b) \textit{Close Grill},  (c) \textit{Open Door}, and (d) \textit{Open Drawer}. The $x$-axis represents translation, and the $y$-axis represents orientation. The origin symbolizes the change in target pose, $\Delta s_{\text{target}}$, while the top right corner denotes the change in source task pose, $\Delta s_{\text{source}}$.}
    \label{fig:landscape}
\end{figure}
% \subsection{Pose Transfer}

%The results of iterations and final loss are reported in table \ref{tab:pose_results}. As is shown in the table, our algorithm has better performance on most tasks. Specifically, our algorithm converges faster with fewer iterations while the direct transfer method tends to fail. Moreover, the final loss of our method is lower than that of direct transfer, which means our method gives a better policy on the transfer target.

The iteration counts $N$ and distances $d$ are detailed in Table \ref{tab:baseline} for \textit{Diff-Transfer}, \textit{MAML}, \textit{DMG}, and \textit{Direct Transfer}. As illustrated in the table, our algorithm manifests superior efficacy across all evaluated tasks. While \textit{MAML} and \textit{DMG} are unable to successfully accomplish any of the four tasks, and \textit{Direct Transfer} only yields a successful outcome in the \textit{Open Door} task, our \textit{Diff-Transfer} manages to fulfill all four tasks, achieving a success rate of $100\%$ across $5$ varied random seeds. Additionally, \textit{Diff-Transfer} requires significantly fewer iterative steps compared to \textit{Direct Transfer} to accomplish the transfer task, underscoring the criticality of constructing a seamless path to mitigate the complexity of each sub-task transfer, and highlighting that attempts to transfer via brute force are frequently either impractical or necessitate more iterations. Regarding \textit{MAML} and \textit{DMG}, these methods, being somewhat antiquated, struggle to finalize this innovative transfer task within a reasonable time.

To confirm the validity of our path-planning approach, we have depicted the landscape of our $Q$-function network in Fig. \ref{fig:landscape}. In each depiction, the horizontal axis denotes the translation, and the vertical axis denotes the orientation, together constituting a task space for any alterations in pose. The origin represents the target pose change $\Delta s_{\text{target}}$ while the top right corner represents the source task pose change $\Delta s_{\text{source}}$. As exhibited in the images, there exists a minimum area surrounding the origin, indicating that the network directs correctly toward the target task. Moreover, this area does not necessarily need to be precisely at the origin; given the varying complexities of different tasks, completing a sub-task pose near the $\Delta s_{\text{source}}$ is often more feasible, resulting in a lower value of $\mathcal{L}_{task}$ in Eq. \ref{eq:l_task} and, subsequently, contributing to a reduced total loss. This task-level characteristic elucidates why these landscapes exhibit a similar pattern with the aforementioned minimum area around the origin, aligning with our anticipations, even though the low-level manipulations might significantly diverge.

\subsection{Ablation Study: \textit{Employ Different Path-Planning Methods}}

We conduct two different ablation tests for \textit{Diff-Transfer} with distinct path-planning methods.

\begin{enumerate}
    \item We remove the Q-learning network and replace it with a deterministic linear interpolation method between $\Delta s_{\text{source}}$ and $\Delta s_{\text{target}}$, denoted as \textit{Linear Interpolation}.
    \item We refine the reward function in Eq. \ref{eq:reward} by removing the task loss term, with $\lambda_t = 0$, denoted as \textit{Diff-Transfer ($\lambda_t=0$)}.
\end{enumerate}

Our experiment results for the ablation study are presented in Table \ref{tab:ablation}. Generally speaking, both \textit{Diff-Transfer} and \textit{Diff-Transfer ($\lambda_t = 0$)} achieve a $100\%$ success rate across four tasks, employing $5$ distinct random seeds, while Linear Interpolation succeeds in three out of the four transfer tasks. This denotes that path planning, even by naive methods, can substantially elevate the success rate in transferring manipulation tasks. To elaborate, the data reveals that our \textit{Diff-Transfer} excels in tasks such as \textit{Close grill}, \textit{Open Door}, and \textit{Open Drawer}, exhibiting quicker convergence (smaller $N$) and heightened precision in manipulation outcomes (smaller $d$) compared to \textit{Diff-Transfer} ($\lambda_t = 0$) and \textit{Linear Interpolation}. Regarding the \textit{Change Clock} task, \textit{Diff-Transfer}, \textit{ablation}, and \textit{Linear Interpolation} display comparable performance, suggesting that accomplishing this transfer task via differentiable physics simulation is relatively uncomplicated. In conclusion, the path-planning methodology employed in \textit{Diff-Transfer} is imperative and efficient, leading to enhanced success rates and reduced time expenditures in most instances.

%% file: tex/con.tex
% This conclusion is written briefly and roughly by yuqi and subject to modification and polishment

In this paper, we introduced an advanced framework aiming to revolutionize the paradigm of robotic manipulation skill acquisition through transfer learning. Drawing inspiration from the omnipresence of Newtonian principles, our method centers on the potential to generalize manipulation strategies across object poses in 3-D Euclidean space. To navigate the complex landscape, we instigate a bridge mechanism, employing a continuum of intermediate sub-tasks as conduits for the seamless relay of skills between distinct object poses, where the path of sub-tasks is generated through a refined $Q$-function network with task-level states and rewards. This focus is further bolstered by our integration of differentiable simulation, affording us an intricate understanding of the physical intricacies inherent in pose transformations. The compelling results from our meticulous experiments underscore the robustness and efficacy of our proposed framework. In summation, our pioneering contributions herald a new era in robotic adaptability, reducing the dependency on ground-up learning and accelerating the skill transfer processes, particularly in the realms of manipulations with different object poses.